\documentclass[lettersize,journal]{IEEEtran}
\usepackage{amsmath,amsfonts}
\usepackage{array}
\usepackage[caption=false,font=normalsize,labelfont=sf,textfont=sf]{subfig}
\usepackage{graphicx}
\usepackage{booktabs}
\usepackage{array}
\usepackage{multirow}
\usepackage{tabularx}
\usepackage{pifont}
\usepackage{makecell}
\usepackage{color}
\usepackage{colortbl}
\usepackage{algorithm}
\usepackage{algpseudocode}

\usepackage{amsmath}
\usepackage{orcidlink}

\usepackage{enumitem}

\newcommand{\mycite}[1]{\textcolor{blue}{\cite{#1}}}
\usepackage{hyperref}
\hypersetup{
    colorlinks=true,    
    citecolor=blue,     
    linkcolor=blue,     
    urlcolor=black       
}

\def\BibTeX{{\rm B\kern-.05em{\sc i\kern-.025em b}\kern-.08em
    T\kern-.1667em\lower.7ex\hbox{E}\kern-.125emX}}
\usepackage{balance}
\begin{document}
\title{fMRI2GES: Co-speech Gesture Reconstruction from fMRI Signal with Dual Brain Decoding Alignment}
\author{Chunzheng Zhu\textsuperscript{\orcidlink{0009-0009-7439-5061}}, 
    Jialin Shao\textsuperscript{\orcidlink{0009-0006-4663-7818}}, 
    Jianxin Lin\textsuperscript{\orcidlink{0000-0003-0359-8821}}, 
    Yijun Wang\textsuperscript{\orcidlink{0000-0002-3372-8167}},
    Jing Wang\textsuperscript{\orcidlink{0000-0003-2734-7138}},
    Jinhui Tang\textsuperscript{\orcidlink{0000-0001-9008-222X}},
    \IEEEmembership{Senior Member, IEEE}
    and Kenli Li\textsuperscript{\orcidlink{0000-0002-2635-7716}}, \IEEEmembership{Senior Member, IEEE}%

\thanks{This research was partially supported by grants from the National Natural Science Foundation of China (Grants No.62472157, No.62202158, No.62206089), the science and technology innovation Program of Hunan Province (Grants No.2023RC3098).}
\thanks{Chunzheng Zhu, Jialin Shao, Jianxin Lin, Yijun Wang, and Kenli Li are with the College of Information Science and Engineering, Hunan University, Changsha, China (e-mail: zhuchzh@hnu.edu.cn; sjlljs176@gmail.com; linjianxin@hnu.edu.cn; wyjun@hnu.edu.cn; lkl@hnu.edu.cn).}
\thanks{Jinhui Tang is with the School of Computer Science and Technology, Nanjing University of Science and Technology, Nanjing 210094, China (e-mail: jinhuitang@njust.edu.cn).}
\thanks{Jing Wang is with the Department of Automation, Tsinghua University,
Beijing 100084, China (e-mail: jwang@njust.edu.cn).}
\thanks{
*Corresponding authors: Jing Wang and Jianxin Lin.
}
}

\markboth{IEEE TRANSACTIONS ON CIRCUITS AND SYSTEMS FOR VIDEO TECHNOLOGY,~Vol.~18, No.~9, September~2020}%
{How to Use the IEEEtran \LaTeX \ Templates}

\maketitle
\begin{abstract}
Understanding how the brain responds to external stimuli and decoding this process has been a significant challenge in neuroscience. While previous studies typically concentrated on brain-to-image and brain-to-language reconstruction, our work strives to reconstruct gestures associated with speech stimuli perceived by brain. Unfortunately, the lack of paired \{brain, speech, gesture\} data hinders the deployment of deep learning models for this purpose.  In this paper, we introduce a novel approach, \textbf{fMRI2GES}, that allows training of fMRI-to-gesture reconstruction networks on unpaired data using \textbf{Dual Brain Decoding Alignment}.  This method relies on two key components: (i) observed texts that elicit brain responses, and (ii) textual descriptions associated with the gestures. Then, instead of training models in a completely supervised manner to find a mapping relationship among the three modalities, we harness an fMRI-to-text model, a text-to-gesture model with paired data and an fMRI-to-gesture model with unpaired data, establishing dual fMRI-to-gesture reconstruction patterns. Afterward, we explicitly align two outputs and train our model in a self-supervision way. We show that our proposed method can reconstruct expressive gestures directly from fMRI recordings. We also investigate fMRI signals from different ROIs in the cortex and how they affect generation results. Overall, we provide new insights into decoding co-speech gestures, thereby advancing our understanding of neuroscience and cognitive science.
\end{abstract}


\section{Introduction}
\label{sec:intro}

\IEEEPARstart{I}{s} it possible to decipher or comprehend human thoughts directly from brain activities? Recent studies \mycite{tang2023semantic,lin2022mind,liu2022learning} have shown the potential to decode various sensory stimuli and mental imagery from brain signals recorded through both non-invasive and invasive methods. This brings us one step closer to the goal of neural decoding.
The current research in this field mainly revolves around two primary directions. The first direction involves reading the perceptual contents encoded in the human brain, such as reconstructing images from brain signals \mycite{beliy2019voxels,lin2022mind,takagi2023high}. The second direction aims to understand cognitive information in the brain, particularly linguistic decoding \mycite{makin2020machine,zou2022cross,proix2022imagined}. This area of study has garnered considerable attention in neuroscience research due to the distinctiveness of human language as an ability that is not shared by other animals. 
An early attempt \mycite{mitchell2008predicting} built a correspondence between brain activation and words by predicting the functional magnetic resonance imaging (fMRI) recordings associated with concrete nouns. Prior to this advancement, our understanding was limited to the discovery that distinct patterns of fMRI activities are associated with images belonging to specific semantic categories \mycite{hanson2004combinatorial, o2005partially, polyn2005category}. After more than a decade of development, the research has progressed towards more complex linguistic decoding tasks, including word decoding \mycite{zou2022cross}, sentence decoding \mycite{makin2020machine} and speech decoding \mycite{moses2019real,proix2022imagined}. However, one type of decoding that has received little attention so far is co-speech gesture. To this point, there have been no attempts made to predict continuous gestures from cortical activities.

\begin{figure}[t]
    \centering
    \resizebox{\linewidth}{!}{
        \includegraphics{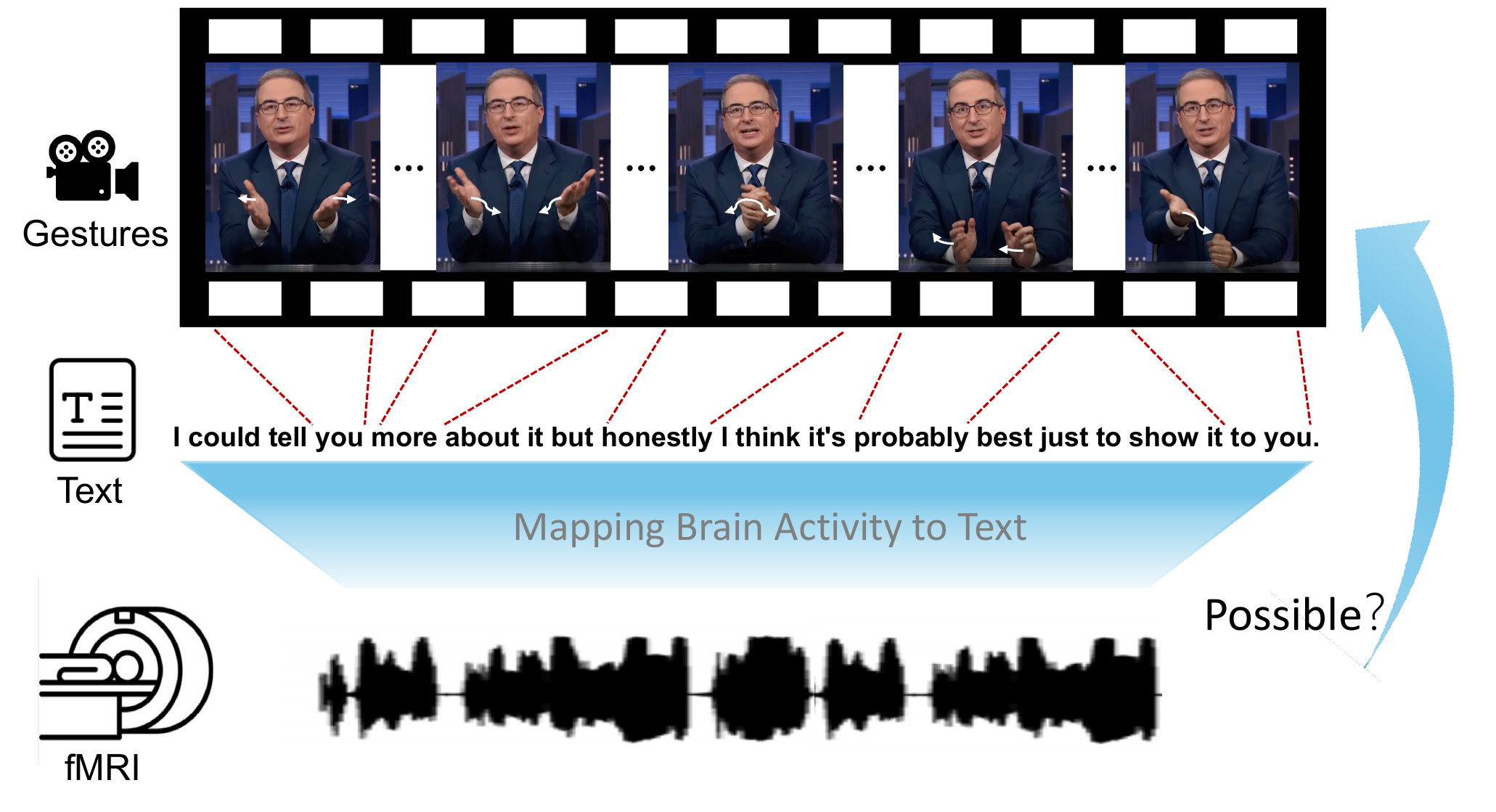}
    }
    \vspace{-4mm}
    \caption{Our work aims to directly reconstruct gestures from brain signals, overcoming the challenge of lacking paired brain, text, and gesture data. We achieve this by leveraging texts that elicit brain activity and corresponding textual descriptions of the gestures.}
    \vspace{-2mm}
    \label{fig:motivation}
\end{figure}

Co-speech gestures commonly occur alongside speech, and these two elements are integral parts of human communication and understanding \mycite{studdert1994hand}. With conscious or unconscious gestures, speakers can better elaborate on the meanings behind the speech content \mycite{cassell1999speech}, \mycite{7208833}, \mycite{9749235}, \mycite{yao2014contour}. Decoding gestures from brain signals is critical to brain-computer interfaces (BCIs), which can help people who are incapacitated or speech-impaired to better communicate with others. It also provides guidance for robots and virtual entities on how to mimic human behavior. Despite its potential, this area has remained unexplored thus far. This is probably attributed to the lack of paired data encompassing brain activation recordings and gestures. Obtaining such data is highly challenging mainly due to the following reasons: 1) Some data collection devices, such as fMRI scanners, typically require subjects to remain as still as possible and avoid any noticeable head or body movements, which contradicts the natural movement in gesture collection. 2) The data collection process is time and labor-consuming. For instance, the NSD dataset \mycite{allen2022massive}, which is widely used for image reconstruction based on fMRI signals, required approximately one year of collection. 3) There is a limited number of volunteers, and only a small fraction of them are willing to participate in invasive recording procedures. 
Additionally, noise in the recorded signals and other factors further complicate the task.

In this paper, we propose to investigate the brain-to-gesture problem and introduce spoken text as an additional modality to compensate for the data deficiency, as depicted in Fig. \ref{fig:motivation}. The rationale for leveraging text modality stems from its intrinsic relationship with speech, while speech provides dynamic and real-time linguistic expressions through acoustic signals, text serves as its static written representation preserving identical semantic content. This speech-text duality establishes a natural bridge for connecting brain activities with gestures. On one hand, gestures accompanying speech are inherently synchronized with the semantic flow of spoken language. On the other hand, the availability of paired {brain, text} and {text, gesture} data allows for the exploration of the relationship between brain activities and gestures through a hybrid approach that combines both supervised and unsupervised learning. Note that our study specially focuses on conversational gestures, namely co-speech gestures, while excluding sign languages (e.g. ASL \mycite{alla1986classifier}) and other types of gestures. Besides, we choose to utilize fMRI data as the representation of the brain modality due to its non-invasive nature, which allows for a wider range of applicability and presents potential future applications \mycite{zou2022cross}.


Based on the analysis mentioned above, we propose a novel approach, fMRI-to-Gesture Reconstruction with Diffusion Models (fMRI2GES), which aims to directly decode gestures from brain recordings. Rather than training a model with supervised brain-gesture data, we leverage existing paired \{fMRI, text\} and \{text, gesture\} data to train an fMRI-to-gesture (F2G) model using unpaired brain-gesture data and dual brain decoding alignment. Our training process comprises two phases. In the initial phase, we construct an fMRI-to-text (F2T) model and a text-to-gesture (T2G) model using supervised learning methods, leveraging corresponding paired data. With these models as a guide, we transition to training the F2G model in an unsupervised manner in the second phase. Given the absence of gesture data for specific fMRI recordings, we develop a novel dual alignment strategy for training. Specifically, we input an fMRI signal into the F2T model to predict a text sequence, which is then passed into the T2G model to generate gestures as pseudo labels. Concurrently, the F2G model also utilizes the fMRI signal as input and produces several gestures.
The F2G model is optimized based on the alignment of these gestures and the pseudo labels, and it stands as the sole required model for inference.

To the best of our knowledge, this study represents the first attempt to decode co-speech gestures from brain activations, thereby addressing a gap in the field of linguistic neural decoding. The main contribution of our work is the proposal of fMRI2GES, a method that enables brain-to-gesture decoding without the need for paired training data of brain signals and gestures. The proposed solution provides novel insights into linguistic decoding, contributing to a deeper understanding of neuroscience. By bridging the gap between brain signals, language, and gestures, this work holds the potential to revolutionize multi-modal communication technologies, offering innovative solutions in human-computer interaction.


In conclusion, our main contributions are summarized as follows:
\begin{itemize}
    \item We introduce a novel fMRI-to-gesture reconstruction method called fMRI2GES, which is the first attempt to decode co-speech gestures from brain activation by utilizing dual brain decoding alignment techniques to directly decode co-speech gestures from brain activity in the absence of paired data.
    
    \item We propose a dual decoding strategy by constructing an fMRI-to-text (F2T) model and a text-to-gesture (T2G) model, employing self-supervised learning to train the fMRI-to-gesture (F2G) model, optimized through generated pseudo labels.

    \item Our experiments demonstrate the establishment of a connection between brain activity and gestures, overcoming the challenge of scarce paired brain-gesture data by incorporating spoken text as a supplementary modality. This research provides new insights into the fields of neuroscience and cognitive science. 
\end{itemize}

\section{Related Works}
\label{sec:related_works}



\begin{figure*}[htbp]
\centering

\begin{minipage}{1\linewidth}
\centering
\includegraphics[scale=0.7]{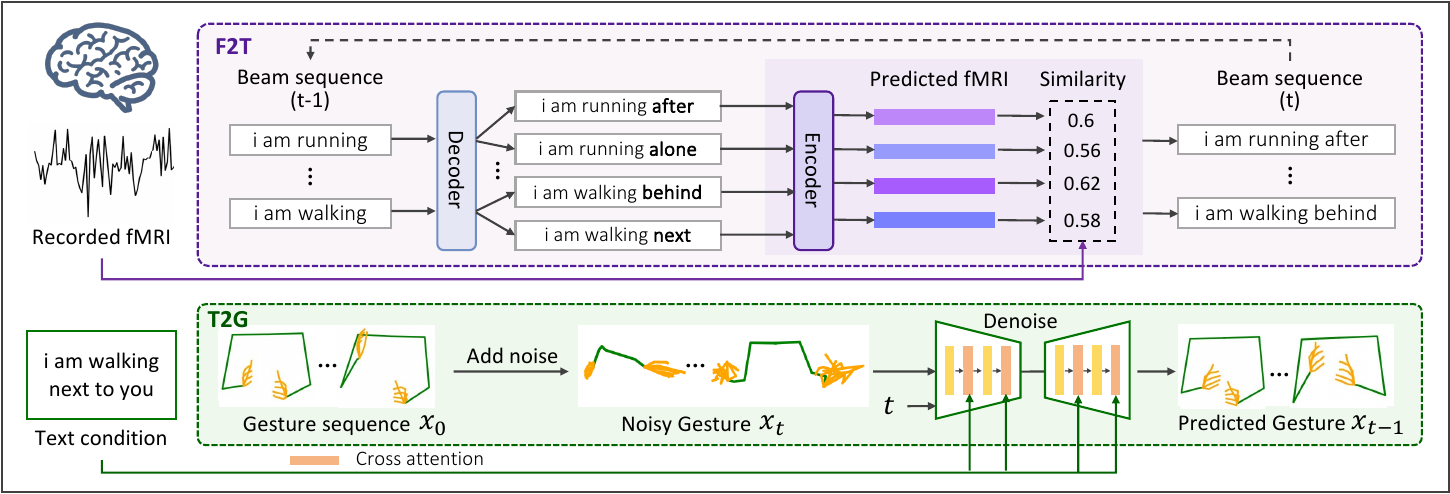}
\footnotesize\text{(a) Training phase I.}\\[2pt]
\vspace{1mm}
\end{minipage}

\vfill

\begin{minipage}{1\linewidth}
\centering
\includegraphics[scale=0.7]{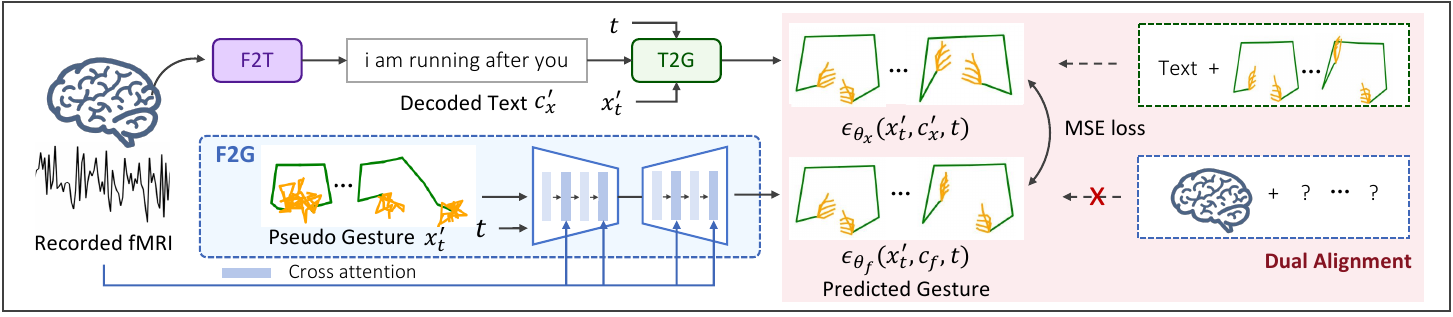}
\footnotesize\text{(b) Training phase II.}\\[2pt]
\end{minipage}

\caption{An overview of our fMRI2GES framework. In the first phase, we conduct supervised training that focuses on learning an fMRI-to-text (F2T) model and a text-to-gesture (T2G) model. In the second phase, we leverage the Dual Brain Decoding Alignment mechanism to train the fMRI-to-gesture (F2G) model in an unsupervised learning manner.}
\label{fig:main}
\vspace{-2mm}
\end{figure*}

\subsection{fMRI-to-Motion.}
Motor imagery, the mental simulation of movement without actual physical execution, has been a subject of significant interest in neuroscience \mycite{hetu2013neural}. Specifically, Schulz et al. conducted research on motor imagery, focusing on reach-to-grasp movements of the right hand in various conditions \mycite{schulz2018action}. Liu et al. further introduced a novel approach to extract fMRI features for discerning between left and right-hand grasping movements, demonstrating the potential of fMRI in decoding motor intentions \mycite{liu2023real}. fMRI signals have also been used to study the effects of upper limb movements and whole body movements on activating cerebral cortical networks \mycite{mizuguchi2014effector, szameitat2007motor}. By monitoring brain activity, researchers can identify movement intentions and translate them into actual movements of the prosthetic limb \mycite{maruishi2004brain, maimon2020artificial}. In the context of rehabilitation, researchers analyze the activity in the brain's motor control areas to design customized rehabilitation programs, aiming to facilitate the recovery of motor abilities in patients \mycite{nunes2023brain, mehler2020graded}. 
However, most existing fMRI-to-Motion works treat the task as a motion classification problem. To the best of our knowledge, there have been no prior studies trying to address the continuous gestures reconstruction problem. 

\subsection{fMRI-to-Speech.}
Decoding speech from brain activity is a long-awaited goal in fields such as healthcare and neuroscience, the initial research on the relationship between the brain and speech was to explore the role of various parts of the brain in decoding speech and to study the relationship between different areas and speech through fMRI \mycite{1998Does,ackermann2007contribution,hickok2003auditory,song2024spiking}. This further pioneered the prediction of brain activation patterns through cross-language semantic decoding \mycite{yang2017commonalities, yang2017commonality}. Related research analyzes brain activity patterns through fMRI, which can spell or classify specific syllables, words or sentences read by subjects. At the same time, the researchers try to analyze the relationship between brain activity and different sound characteristics to decode the speech content heard by the subject \mycite{buchweitz2012identifying,otaka2008decoding,mitchell2008predicting,pasley2012reconstructing,correia2015decoding}. Compared to speech decoders for intrusive recordings, speech decoders for non-intrusive recordings can be used more widely. Tang et al. \mycite{tang2023semantic}  created a non-invasive language decoder, used fMRI data to build a non-invasive brain-computer interface, and translated thoughts into continuous text or speech by decoding brain activity, proving the feasibility of a non-invasive language brain-computer interface.
All these works collectively underscore the richness of information embedded within fMRI signals and the exciting potential to extend decoding capabilities to include co-speech gestures, a critical aspect of human communication.

\subsection{fMRI-to-Image.}
Recently, the reconstruction of images from fMRI data has gained significant attention, thanks to the rapid advancements of deep learning techniques. Traditional fMRI signal decoding involves manual fitting of fMRI data to generate image features, but the outcomes of these methods tend to be blurred and depend on manual configuration \mycite{naselaris2009bayesian, kay2008identifying, fujiwara2013modular}. With deep learning, some studies \mycite{beliy2019voxels, gaziv2022self, chapelle2009semi} use semi-supervised learning to train the encoder-decoder structure to solve the problem of insufficient matching pairs for fMRI stimulus reconstruction. Many recent studies  \mycite{lin2022mind, ma2025brainclip, chen2024disendreamer} on reconstructing natural images have begun to introduce text information, such as combining multi-modal pre-training models CLIP \mycite{radford2021learning}. Other works \mycite{takagi2023high,lu2023minddiffuser,huo2024neuropictor,10649652} also seek to leverage the powerful Diffusion Models \mycite{ho2020denoising,rombach2022high} to reconstruct high-resolution images.
The advancements in fMRI-to-Image techniques have been an inspiration for our approach to explore the application of diffusion models for fMRI-to-Gesture reconstruction.

\subsection{Co-speech Gesture Generation.}
For traditional Co-speech Gesture Generation, various rule-based approaches \mycite{cassell1994animated, lee2006nonverbal, huang2012robot, marsella2013virtual, lhommet2015cerebella} were adopted, and recent research has turned to neural networks to achieve mapping to gestures in an end-to-end manner. These models employ various neural network architectures, including CNN \mycite{habibie2021learning}, RNN \mycite{yoon2019robots}, and Transformer \mycite{bhattacharya2021text2gestures}, to achieve more natural gesture generation. To improve the fidelity and expressiveness of gesture generation, researchers have introduced probabilistic modeling techniques such as VAEs \mycite{ghorbani2023zeroeggs}, VQ-VAEs \mycite{liu2022audio, yazdian2022gesture2vec}, normalized flow \mycite{alexanderson2020style}, and GANs \mycite{ginosar2019learning, qian2021speech}. Diffusion models also show potential for capturing complex gesture distributions and have produced impressive performance in co-speech gesture generation \mycite{petrovich2023tmr, zhu2023taming, cheng2024siggesture, chen2024diffsheg}. Based on this, we further explore the diffusion model to capture gesture-related information in fMRI signals extracted from complex brain activities.

\section{Method}

In this section, we present a novel fMRI-to-Gesture Reconstruction with Diffusion Models (fMRI2GES) framework for directly decoding gestures from fMRI signals. Since there is no paired \{fMRI, gesture\} training corpus, we attempt to solve the mapping between fMRI recordings and their corresponding co-speech gestures, using two paired \{fMRI, text\} and \{text, gesture\} datasets. 
Our training process consists of two phases, as illustrated in Fig. \ref{fig:main}. In the first phase, 
we conduct supervised training that focuses on the transition from fMRI to text and from text to gesture, learning an fMRI-to-text (F2T) model and a text-to-gesture (T2G) model.
In the second phase, 
we shift our focus to training the fMRI-to-gesture (F2G) model, leveraging the F2T and T2G models obtained during the first phase. This two-phase training strategy, combining both supervised and unsupervised learning, enables us to harness the power of both labeled and unlabeled data for effective training and improved model performance. It is worth noting that the data across all modalities remain temporally aligned during the reverse process, ensuring that the sequence \(c_x^{\prime}\), obtained by F2T, has been preprocessed and encoded through duplication. Furthermore, the gesture sequence \(x_t^{\prime}\) is aligned with the fMRI signals from the relevant brain regions utilized in various experiments during data preprocessing, as detailed in \ref{datapreprocess}.


\subsection{Problem Formulation}
\textbf{Notation. }
Given a record of fMRI, the task of fMRI-to-gesture reconstruction is to predict a sequence of gestures $x$ that is compliant to the semantic information of the stimuli which evokes the brain activity, where $c_f \in \mathbb{R}^{N \times D_f}$ is $N$ segments of fMRI signals, each consisting of $D_f$ voxels, $x \in \mathbb{R}^{N \times 98}$ is a sequence of $N$ gestures with 49 2D-keypoint coordinates. The output of the F2T model and the input of the T2G model are sequences of utterance words, with the embedding denoted as $c_x \in \mathbb{R}^{N \times D_x}$.
We ensure that the lengths of the words, fMRI signals and gesture frames are identical. To achieve this, we replicate a word and a segment of fMRI recording when they are related to multiple frames in the original data \mycite{asakawa2022evaluation}. 
In light of the foundation of our T2G and F2G models on diffusion models, we begin by explaining how to execute gesture generation using denoising diffusion probabilistic models (DDPMs) \mycite{ho2020denoising}. The diffusion model encompasses two distinct processes, namely, a forward process and a reverse process.

\textbf{Forward Process.} 
In the forward diffusion process, Gaussian noise is progressively introduced to the source gesture sequence $x_0$, which is sampled from data distribution $q(x_0)$. Given a variance schedule $\beta_1, \ldots, \beta_T$, the forward process adheres to a Markov chain. For $t\in(0,T]$, the state transition from $x_{t-1}$ to $x_{t}$ is calculated as:
\begin{equation}
q(x_t|x_{t-1}) := \mathcal{N}(x_t; \sqrt{1 - \beta_t}x_{t-1}, \beta_t \mathbf{I}).
\end{equation}
The forward process provides an estimation of the posterior distribution $q(x_{1:T}|x_0)=\prod_{t=1}^T q(x_t|x_{t-1})$, where $x_{1:T}$ represents latent variables with dimensions matching $x_0$. 

\textbf{Reverse Process.} 
The reverse process, denoted as $p_\theta(x_{0:T})$, aims to restore the original gesture sequence. This process also follows a Markovian scheme, featuring learned Gaussian transitions that initiate from $x_T \sim \mathcal{N}(0, \mathbf{I})$. The reverse operation of the continuous diffusion process maintains an identical transition distribution structure \mycite{ho2020denoising}. This insight encourages us to employ a Gaussian transition for modeling $p_{\theta}(x_{t-1}|x_{t})$ within an unconditional context:
\begin{equation}\label{eq:denoise}
p_{\theta}(x_{t-1}|x_{t}) := \mathcal{N}(x_{t-1}; \mu_{\theta}(x_t,t), \Sigma_{\theta}(x_t, t)),
\end{equation}
where the corrupted noisy data \(x_t\) is sampled by \(q(x_t|x_0) = \mathcal{N}\left(x_t; \sqrt{\bar{\alpha}_t} x_0, (1 - \bar{\alpha}_t)\mathbf{I}\right)\), \(\alpha_t = 1 - \beta_t\) and \(\bar{\alpha}_t = \prod_{s=1}^t \alpha_s\).

\textbf{Training  Objective.} The whole diffusion network can be optimized with the loss function:
\begin{equation}
\small
\mathcal{L}(\theta) = \mathbb{E}_q\left[\sum_{t=2}^T D_{KL}(q(x_{t-1}|x_t, x_0)||p_{\theta}(x_{t-1}|x_t))\right].
\end{equation}
Through reparameterization \mycite{kingma2013auto}, we can further simplify the training objective to the form of MSE loss as:
\begin{equation}\label{eqa:mse}
\mathcal{L}(\theta) = \mathbb{E}_{\epsilon,t} \|\epsilon - \epsilon_\theta(x_t, t)\|_2^2.
\end{equation}

\subsection{Decoding text from fMRI Signals}

As is well known, fMRI uses the blood-oxygen-level-dependent (BOLD) signal to capture brain images by tracing the changes in the blood flow. A neural spike results in approximately 10 seconds of fluctuation in the BOLD signal, which means that fMRI has a naturally low temporal resolution \mycite{tang2023semantic}. In the same 10 seconds, a native English speaker can speak around 20 words, which is significantly more than the number of brain images. This presents an ill-posed problem for fMRI-to-text decoding. 


To address the aforementioned issue, we develop the F2T model based on the approach proposed in \mycite{tang2023semantic}, utilizing the strategy of envisioning candidate sequences. To enhance contextual understanding and achieve more natural language decoding, we integrate GPT-2 \mycite{radford2019language} with a kernel sampling strategy.
By leveraging GPT-2 to predict multiple possible word sequences, the probabilistic distribution of the language model compensates for insufficient temporal sampling. The candidate sequences are then filtered by an encoding model to select the optimal solution. Specifically, the F2T model consists of a decoder and an encoder, as depicted in Fig. \ref{fig:main}\textcolor{blue}{(a)}. The decoder aims to efficiently decode text sequences from fMRI signals, while the encoder evaluates the candidate sequences by predicting brain activities given these sequences.
Technically, given fMRI signal $c_f \in \mathbb{R}^{N \times D_f}$ containing $N$ segments with $D_f$ voxels, we formulate the decoding task using Bayes' theorem as:
\begin{equation}
\hat{w}_{1:M} = \mathop{\text{argmax}}_{w_{1:M}} \underbrace{p_\theta(w_{1:M}|c_f)}_{\text{Posterior}} \propto \underbrace{p_\text{LM}(w_{1:M})}_{\text{Language prior}} \cdot \underbrace{p_\phi(c_f|w_{1:M})}_{\text{Brain likelihood}},
\end{equation}
where $\theta$ denotes decoder parameters for text generation, $\phi$ indicates encoder parameters for brain activity prediction, the language prior  $p_{\text{LM}}(w_{1:M})$  ensures fluency using GPT-2, and $w_{1:M}$ denotes the decoded word sequence. To align the low-resolution fMRI segments with high-rate word sequences, we partition $c_f$ into overlapping windows via dynamic time warping, assigning each window to a subsequence of words.

The decoder generates candidate sequences using nucleus sampling \mycite{holtzman2019curious}, which adaptively controls diversity through probability thresholding. For each time step $t$, we first sort the token probabilities in descending order, denoted as $P(w^{(1)}_t|w_{1:t-1}) \geq P(w^{(2)}_t|w_{1:t-1}) \geq \dots$, and define the nucleus set as:
\begin{equation}
\text{Top-}p(w_t) = \min\left\{k \in \mathbb{N} \big| \sum_{i=1}^k P(w^{(i)}_t|w_{1:t-1}) \geq 0.9\right\}.
\end{equation}

From this set, we sample candidate extensions for each sequence in the beam. At each time step, the encoder computes the similarity between the predicted $\hat{c}_f$ and the observed $c_f$ to score candidate sequences. The top-$k$ highest-scoring candidates are retained in the beam, dynamically adjusting $k$ based on the entropy of the language model predictions. This iterative process progressively refines the decoded sequence $c_x^{\prime}$ from $c_f$, initializing with an empty sequence and contextualizing via previously decoded words.

\subsection{Text-conditioned Gesture Generation}\label{sec:3.3}
Recently, it has been verified that diffusion models, such as Stable Diffusion \mycite{rombach2022high}, have the capability to model conditional distributions $p_\theta(x_0|c)$ over conditions $c$ by extending the UNet backbone with a condition-specific encoder that employs a cross-attention mechanism \mycite{vaswani2017attention}.
Motivated by this, we develop our T2G model based on conditional diffusion models with noise-free text inputs to eliminate semantic-irrelevant acoustic artifacts.  To better capture the interactions between modalities, we leverage cross-attention mechanisms to investigate and model the cross-modal relationship, ensuring a more robust alignment between textual and gestural representations.

Technically, given paired text sequence $c_x$ and corresponding gesture sequence $x_0$ sampled from real dataset, our goal is to learn a model distribution $p_{\theta_x}(x_0)$ parameterized by $\theta_x$ that approximates real gesture distribution $q(x_0)$ conditioned on $c_x$, which is realized by iteratively denoising the noised latent state $x_t$ to the output gesture sequence $x_0$.
According to Equation \ref{eq:denoise}, the reverse process of each time step can be rewritten as
\begin{equation}\label{eq:denoise_text}
\small
p_{\theta}(x_{t-1}|x_{t},c_x) := \mathcal{N}(x_{t-1}; \mu_{\theta_x}(x_t,t,c_x), \Sigma_{\theta_x}(x_t, t,c_x)).
\end{equation}
To specifically incorporate the condition into the reverse process, we use a cross-attention mechanism to map $c_x$ to the intermediate layers of the UNet $\theta_x$:
\begin{equation}\label{eq:attention}
\text{Attention}(Q, K, V) = \text{softmax}\left(\frac{QK^T}{\sqrt{d}}\right) \cdot V, 
\end{equation}
where \(Q = W^{(x)}_Q \cdot z_t\), \(K = W^{(x)}_K \cdot c_x\), and \(V = W^{(x)}_V \cdot c_x\), $z_t$ represents a (flattened) intermediate representation of the UNet implementing,  
\(W^{(x)}_V \), \(W^{(x)}_Q\), and \(W^{(x)}_K \) are learnable projection matrices. We achieve deep fusion of text sequences and gesture latent states. The network can be optimized using the same loss function conditioned on context $c_x$, as specified in Equation \ref{eqa:mse}:
\begin{equation}
\mathcal{L}_{t2g}(\theta_x) = \mathbb{E}_{\epsilon,t} \|\epsilon - \epsilon_{\theta_x}(x_t,c_x, t)\|_2^2.
\end{equation}

\subsection{fMRI-conditioned Gesture Reconstruction}
The F2G model shares a similar structure with the T2G model introduced in Section \ref{sec:3.3}. In this case, the condition is transitioned to the recorded fMRI signals. Our objective is to learn the model distribution $p_{\theta_f}(x_0)$, parameterized by $\theta_f$, which approximates the real gesture distribution $q(x_0)$ conditioned on the fMRI sequence $c_f$. F2G also employs a cross-attention mechanism to map the fMRI sequence to the intermediate layers of the UNet, as elaborated in Equation \ref{eq:attention}. 
To enable the training of F2G, we develop the dual brain decoding alignment mechanism, which relies on the pseudo-labels generated through the cascaded F2T and T2G models.
In addition, there naturally exists a domain gap between the decoded text sequence $c_x^{\prime}$ (mainly about stories) from fMRI signal and spoken text $c_x$ from real data (mainly about television shows or
university lectures). By using dual brain decoding alignment, we can narrow down such domain gap by fine-tuning the T2G model under both the new and existing constraints.

\begin{algorithm}[t]
\caption{Training Process for fMRI-conditioned Gesture Reconstruction}
\label{alg:training}
\begin{algorithmic}[1]
\Repeat
    \State Sample $(x_0, c_x) \sim q(x_0, c_x)$, $c_f \sim q(c_f)$
    \State $c_x^{\prime} \gets \text{F2T}(c_f)$
    \State Sample $t \sim \text{Uniform}(\{1, \ldots, T\})$ 
    \State Sample $x_t^{\prime}$ from T2G($c_x^{\prime}$)
    \State Compute $\nabla_{\theta_x,\theta_f} \|\epsilon - \epsilon_{\theta_x}(x_t,c_x, t)\|_2^2 +$
    \Statex \quad \quad \quad  \quad $ \lambda\sqrt{\frac{1 - \bar{\alpha}_t}{\bar{\alpha}_t}}\left\| \epsilon_{\theta_x}(x_t^{\prime}, c_x^{\prime}, t)-\epsilon_{\theta_f}(x_t^{\prime}, c_f, t) \right\|_2^2 $ 
    \State Perform gradient descent 
\Until{converged}
\end{algorithmic}
\end{algorithm}
\textbf{Dual Brain Decoding Alignment.}
Considering the disparity in distributions between different datasets, we introduce a dual-way decoding alignment mechanism for precise semantic transfer and employ a conditional diffusion model to bridge the gap between the two distributions by conditioning gesture generation on textual descriptions and fMRI signals. In the first branch, given the fMRI signal $c_{f}$, we first obtain the approximated text sequence $c_x^{\prime}$ using the F2T model. Then, a pseudo $x_t^{\prime}$ can be obtained by repeatedly sampling from the reverse process of T2G, as described in Equation \ref{eq:denoise_text} in the model $\theta_x$.
We assume a likelihood function $P(x_0, c_{f})$ that represents the gesture $x_0$ falling within the data distribution specified by the fMRI condition $c_{f}$. If $P(x_0, c_{f})$ is close to 1, this indicates that the gesture condition $x_0$ precisely fits the data distribution described by the fMRI condition $c_{f}$. Given this assumption, we would expect that $G_f(x_t^{\prime}, c_{f},x_0) = x_0$ for any noised latent $x_t^{\prime}$.
We optimize the F2G model by minimizing the MSE loss of the generated gesture and the pseudo gesture:

\begin{small}
\begin{align}
\label{eq:10}
\mathcal{L}_{dual}(\theta_x,\theta_f) = & E_{x_t^{\prime},\epsilon,t} \left\| x_0- G_f(x_t^{\prime}, c_{f}, x_0) \right\|_2^2 \nonumber \\ 
= & E_{x_t^{\prime},\epsilon,t} \left\| G_x(x_t^{\prime}, c_x^{\prime}, x_0) - G_f(x_t^{\prime}, c_f, x_0) \right\|_2^2 \nonumber \\ 
= & E_{x_t^{\prime},\epsilon,t} \left\| \frac{1}{\sqrt{\bar{\alpha}_t}} \left( x_t^{\prime} -\sqrt{1 - \bar{\alpha}_t}  \epsilon_{\theta_x}(x_t^{\prime}, c_x^{\prime}, t) \right)- \right. \nonumber \\
& \left. \frac{1}{\sqrt{\bar{\alpha}_t}} \left(x_t^{\prime}- \sqrt{1 - \bar{\alpha}_t}  \epsilon_{\theta_f}(x_t^{\prime}, c_f, t) \right) \right\|_2^2 \nonumber \\
= & E_{x_t^{\prime},\epsilon,t} \sqrt{\frac{1 - \bar{\alpha}_t}{\bar{\alpha}_t}}\left\| \epsilon_{\theta_x}(x_t^{\prime}, c_x^{\prime}, t)-\epsilon_{\theta_f}(x_t^{\prime}, c_f, t) \right\|_2^2,
\end{align}
\end{small}
where $x_0$ could be directly predicted from $x_t$ using DDIM sampling \mycite{songdenoising}:
\begin{equation}\label{eq:zt2z0}
x_0 = \frac{x_t}{ \sqrt{\bar{\alpha}_t}} - \frac{\sqrt{1 - \bar{\alpha}_t} \epsilon_\theta(x_t, c, t)}{\sqrt{\bar{\alpha}_t}}.
\end{equation}

The pseudocode for the training process can be found in Algorithm \ref{alg:training}.




\begin{table*}[t]
\centering
\renewcommand{\arraystretch}{1.5}
\setlist[itemize]{nosep, leftmargin=*, before=\vspace{-0.5\baselineskip}, after=\vspace{-\baselineskip}}
\caption{Advantages of F2T, T2G and F2G models over existing methods.}
\begin{tabularx}{\textwidth}{>{\raggedright\arraybackslash}p{2.5cm} *{3}{>{\centering\arraybackslash}X}}
\toprule
\textbf{Procedures} & \hspace{-0.7cm}\textbf{Baseline} & \hspace{-0.2cm}\textbf{fMRI2GES (ours)} & \hspace{-0.2cm}\textbf{Improvement}  \\
\midrule
\vspace{0.03cm}
\textbf{fMRI to Text}
& \begin{itemize}
  \item GPT-1 for language decoding
  \item Deterministic Beam Search for candidate sequences
  \end{itemize}
& \begin{itemize}
  \item GPT-2 for language decoding
  \item Kernel sampling to stochastically explore candidate sequences
  \end{itemize}
& \begin{itemize}
  \item Stronger contextual understanding
  \item Ensuring the decoding of more natural and diverse sequences
  \end{itemize} \\ 
 \addlinespace
 \hline
\addlinespace
\vspace{0.03cm}
\textbf{Text to Gesture}
& \begin{itemize}
  \item Noisy speech driven generation
  \item Weakly conditional KL constraint
  \item Self-attention mechanism
  \end{itemize}
& \begin{itemize}
  \item Text-aligned noise-free generation
  \item Conditional MSE loss
  \item Cross-attention interaction
  \end{itemize}
& \begin{itemize}
  \item Semantically guided accurate gestures
  \item Text-gesture tight coupling
  \item Enhanced cross-modal generation
  \end{itemize} \\ \addlinespace
 \hline
\addlinespace
\vspace{-0.01cm}
\textbf{fMRI to Gesture}
& \begin{itemize}
  \item[--] \rule{0pt}{1.5\normalbaselineskip} None
  \end{itemize}
& \begin{itemize}
  \item Dual brain decoding alignment
  \item Cross-modal latent space fusion
  \item Pseudo-label guided generation
  \end{itemize}
& \begin{itemize}
  \item First fMRI-to-Gesture work
  \item Domain gap elimination
  \item Training in an unsupervised manner
  \end{itemize} \\
  \addlinespace
\bottomrule
\end{tabularx}
\label{tab:pipeline}
\end{table*}

\section{Experiments}
\subsection{Datasets and Metrics}
\label{Datasets and settings}

\subsubsection{fMRI-to-Text Dataset (F2TD)} This dataset provides paired \{fMRI, text\} data collected by Tang et al. \mycite{tang2023semantic}. It contains fMRI recordings collected from seven subjects, comprising three female and four male participants, using a 3T Siemens Skyra scanner. The participants were asked to receive auditory stimuli and create narratives without speaking, while their brain activities were recorded simultaneously. Our approach posits that the information from different subjects responding to the same stimuli indicates that the brain activity of one subject can be linearly expressed by that of another \mycite{ferrante2024towards}. This allows for the mitigation of individual differences during ridge regression, resulting in similar decoding outcomes across different subjects. The experiments are primarily on voxel data from the auditory cortex and several speech-related areas, which are closely linked to human language capabilities.

\subsubsection{Text-to-Gesture Dataset (T2GD)}

The Specific Gesture Dataset \mycite{ginosar2019learning} is a comprehensive video dataset specifically collected for the study of speech and gesture. It consists of an extensive collection of 144 hours of videos featuring 10 gesturing speakers. The gestures performed by the speakers are represented by 49 2D skeletal keypoints, including key positions such as the neck, shoulders, elbows, wrists, and hands, which are detected using OpenPose \mycite{cao2017realtime}. Additionally, Asakawa et al. \mycite{asakawa2022evaluation} enhanced the dataset by incorporating text information into the videos through the use of the Cloud Speech-to-Text service provided by Google Cloud. This process further constructed the paired T2GD utilized in this paper.


\subsubsection{Evaluation Metrics}
Following \mycite{asakawa2022evaluation}, we quantitatively measured the generated gestures by computing the errors between the generated gestures and the reference gestures. Since we did not have ground-truth gestures corresponding to fMRI signals, we compared our generated gestures with the gestures decoded from the corresponding text, verifying the potential for fMRI-generated gestures. The mean absolute error (MAE) and average position error (APE) were adopted for this purpose. Intuitively, smaller MAE and APE values indicate that the generated gestures are much closer to pseudo gestures. 
Additionally, we also calculated the Probability of Correct Keypoints (PCK) \mycite{yang2012articulated} and the Fr\'{e}chet gesture distance (FGD) \mycite{yoon2020speech} of gestures. PCK is a metric of pose detection and we used it to measure whether the keypoints of the gestures generated by F2G and T2G are within a distance threshold. FGD is often used for image generation measurement. We employed it to measure the difference between the two gesture distributions.
Lastly, We utilize the Beat Consistency Score (BC) \mycite{li2022danceformer, li2021ai} to quantify the synchronicity of rhythms and Diversity \mycite{NEURIPS2019_7ca57a9f} to evaluate the range of distinctions in gestures generated from different inputs.

\subsection{Dataset Preprocessing Details}  

\label{datapreprocess}

One speaker can speak many words and express his thoughts with many gestures in a period of time \mycite{kita2017gestures}. In fact, the number of words and the number of gestures are hardly equal to each other, causing a many-to-many relationship between words and gestures. Hence, we take into account the occurrence time of each word from each video and align the words to gesture frames at 15fps by duplicating these words, thereby constructing the paired \{text, gesture\} data from Text-to-Gesture Dataset (T2GD). The T2G model in phase I is trained on the aligned T2GD.
For the voxel-wise model, an fMRI response often corresponds to multiple-word stimuli within a repetition time (TR), where TR is 2s. Considering that the text sequences and gesture sequences are strictly aligned at every time point, the one-to-many relationship between an fMRI signal and multiple-word stimuli requires us to align the fMRI signals to the text sequences. We use the F2T model to generate a text sequence from an fMRI signal, and words of the text sequence are duplicated following the same strategy of T2GD. The duplicated text sequences are used for the T2G model in phase II.


Specifically, for fMRI signal $\mathbf{c}_f$ within a TR interval and 15 fps gesture videos with 2D keypoints, each fMRI segment $\mathbf{c}_f^{(i)}$ corresponds to a 2s video clip containing $N = \lceil 15 \times 2 \rceil = 30$ gesture frames. The decoded words $\{\mathbf{c}_x^{\prime (w)}\}_{w=1}^W$ are  replicated $\eta_w = \lceil N/W \rceil$ times to align temporal dimensions with the indicator $\mathbf{1}_{\eta_w}$, where $W$ denotes the number of words per TR. The alignment is formalized as:
\begin{equation}
    \mathbf{c}_f = \underbrace{\left[ \mathbf{c}_f^{(i)}, \dots, \mathbf{c}_f^{(i)} \right]}_{N \text{ replicates }}, \quad 
    \mathbf{c}_x^{\prime} = \underbrace{\left[\bigoplus_{w=1}^W \, \underbrace{\left( \mathbf{c}_x^{\prime (w)} \otimes \mathbf{1}_{\eta_w} \right)}_{\substack{\text{Replicate $\eta_w$ times}}}\right]}_{ \\ \sum_{w=1}^W \eta_w = N \text{ replicates }}.
\end{equation}
Our approach simplifies the alignment by the process above, eliminating the need for high-frequency fMRI sampling. This is addressed by upsampling to compensate for information loss, and we have conducted specific experiments to validate its feasibility. Currently, our method remains straightforward and exploratory. Future researchers could explore the integration of neural lag adjustments, which may enhance the temporal alignment between fMRI signals and gesture data, ultimately leading to more accurate reconstructions.

\subsection{Implementation details} 

In phase I, two models, F2T and T2G, are trained through a supervised learning method with paired {fMRI, text} and {text, gesture}data respectively. The F2T encoder establishes a correlation between the fMRI signals and the semantic information of the stimulus words. 
\label{A.3.1}
Each of these stimulus words is transitioned to a 768-dimensional semantic embedding through a GPT-2 model \mycite{radford2019language}. An initial stimulus is constructed via a three-lobe Lanczos resampling and 4-TR delays initial stimuli are concatenated to construct the final 3072-dimensional stimulus to predict an fMRI signal $C_f$ by the F2T encoder at time \textit{t}.
Before decoding, a word rate model is established to predict the word rate \textit{w} from the corresponding fMRI signal $C_f$ through the weights estimated by regularized linear regression. When the word rate model detects new words at time \textit{t}, the language model GPT-2 generates the next possible words with a beam width \textit{k} for each of the latest candidates. The F2T decoder maintains a beam of \textit{k} candidates consistently that present the most likely word sequences at time \textit{t}, with the beam search \mycite{tillmann2003word} and nucleus sampling \mycite{binder2009semantic} algorithms.
In the T2G stage, the text sequence is encoded by FastText \mycite{bojanowski2017enriching} as a text condition, and the text input vectors are processed into ${N \times 300}$ uniform-sized features. We propose the T2G model based on the conditional diffusion model and use the cross-attention mechanism for cross-modality study. Specifically, the goal of the model is to obtain an approximately true gesture distribution $q(x_0)$ based on a given paired text sequence $c_x$.


In phase II, the F2G model is trained with the fMRI data via an unsupervised learning approach that combines the F2T and T2G models. To effectively narrow the domain gap between the decoded text sequence \(c_x^{\prime}\) derived from fMRI and the actual spoken text, it is crucial to decode and align the gestures generated by the T2G and F2G models at both ends of the process.
The first branch utilizes the pre-trained T2G model from phase I to generate a gesture sequence \(x_t^{\prime}\) based on the duplicated decoded text sequence \(c_x^{\prime}\) obtained from the fMRI signal through the F2T model, using \(c_x^{\prime}\) as a textual condition. The second branch, F2G, employs a similar structure to T2G but conditions on the fMRI signal \(c_f\). In this branch, for each block of the Unet architecture, sinusoidal absolute positional encoding of the time step \(t\) is added to the input. The optimization objective involves minimizing the discrepancy between the generated gestures and the pseudo gestures through MSE loss, thereby achieving dual-way alignment. The improvements of our approach compared to previous methods are summarized in Table \ref{tab:pipeline}.

\begin{table*}[t]
\centering
\caption{Quantitative evaluations (MAE, APE, PCK, FGD, BC, and Diversity) of the gestures generated using our F2G model. Pseudo GT gestures are generated by the T2G model.}
\label{tab:1}
\setlength{\tabcolsep}{8pt}
\begin{tabular}{p{39pt} p{49pt} p{49pt} p{49pt} p{49pt} p{49pt} p{39pt} p{43pt}<{\centering}}
\toprule
\multirow{2}{*}{Model} & \multirow{2}{*}{Signal} & \multicolumn{6}{c}{Evaluation Metrics} \\
\cmidrule(lr){3-8}
& & MAE $\downarrow$ & APE $\downarrow$ & PCK $\uparrow$ & FGD $\downarrow$ & BC $\uparrow$ & Diversity $\uparrow$ \\
\midrule
\multirow{2}{*}{F2G} & Noise & 0.929 & 1.453 & 0.206 & \textbf{16.0} & 0.338 & 90.19\\
 & fMRI & \textbf{0.603} & \textbf{0.874} & \textbf{0.451} & 25.1 & \textbf{0.628} & \textbf{163.7} \\
\bottomrule
\end{tabular}
\end{table*}

\begin{table*}[tb]
\centering
\caption{Comparison of results using different fMRI conditions from various cortical regions. 'S+A' represents Speech + Auditory, while 'Motor' refers to the motion control cortex.}
\label{tab:2}
\setlength{\tabcolsep}{8pt}
\begin{tabular}{p{39pt} p{49pt} p{49pt} p{49pt} p{49pt} p{49pt} p{39pt} p{43pt}<{\centering}}
\toprule
\multirow{2}{*}{Model} & \multirow{2}{*}{Region} & \multicolumn{6}{c}{Evaluation Metrics} \\
\cmidrule(lr){3-8}
& & MAE $\downarrow$ & APE $\downarrow$ & PCK $\uparrow$ & FGD $\downarrow$ & BC $\uparrow$ & Diversity $\uparrow$ \\
\midrule
\multirow{4}{*}{F2G}  & All & 1.248 & 1.800 & 0.215 & 63.4 & 0.341 & 92.25\\
  & S+A & 0.607 & 0.911 & 0.444 & 24.5 & 0.619 & 164.8\\
  & Speech & 0.621 & 1.029 & 0.439 & \textbf{24.4} & 0.615 & \textbf{166.3} \\
  & Auditory & \textbf{0.603} & \textbf{0.874} & \textbf{0.451} & 25.1 & \textbf{0.628} & 163.7 \\
   & Motor & 1.182 & 1.652 & 0.225 & 61.8 & 0.358 & 95.4 \\
\bottomrule
\end{tabular}
\end{table*}

\begin{figure*}[t]
    \centering
    \resizebox{0.99\linewidth}{!}{
        \includegraphics{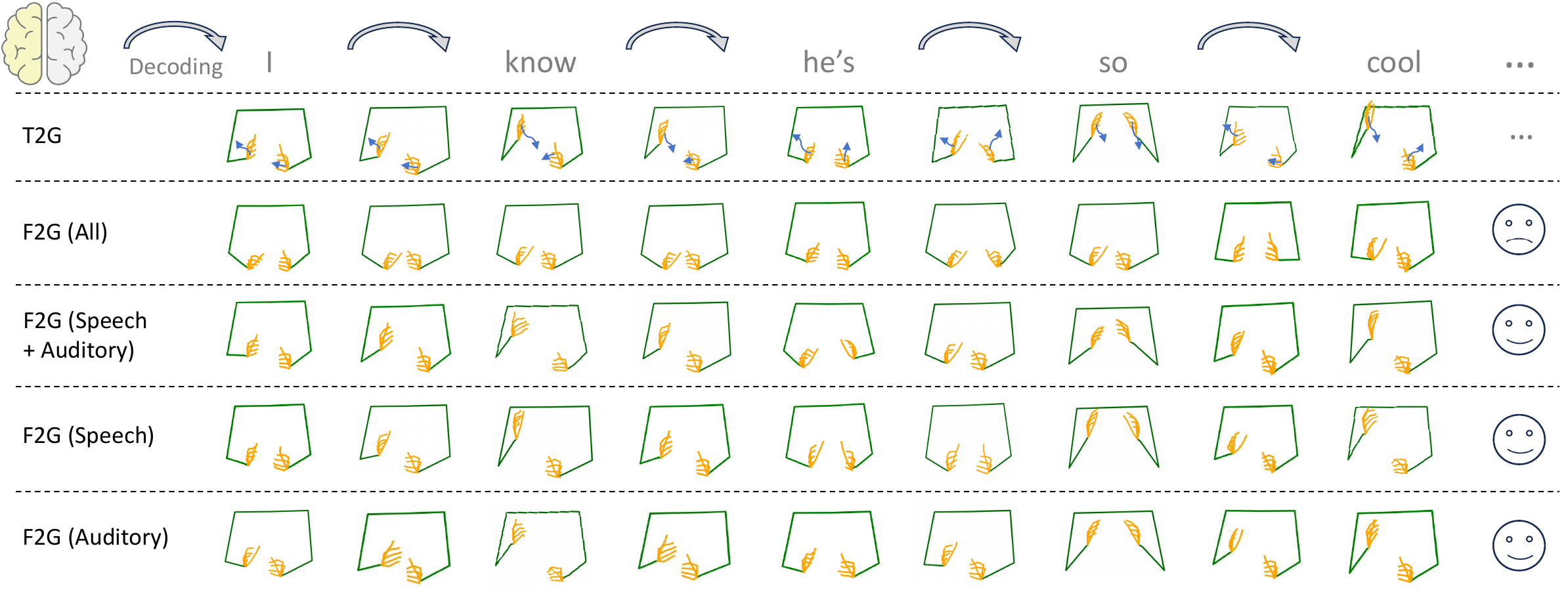}
    }
    \caption{Gesture examples were generated using T2G and F2G. F2G (All) utilizes voxel-sampled fMRI data from the entire brain, F2G (Auditory) focuses on auditory regions, and F2G (Speech) focuses on speech-related areas. Additionally, F2G (S+A) integrates fMRI data from both language and auditory regions.}
    \vspace{-2mm}
    \label{fig:gesture}
\end{figure*}

\begin{table*}[t]
\centering
\caption{Effect of different noise levels on various evaluation metrics.}
\label{tab:noise}
\setlength{\tabcolsep}{8pt} 
\begin{tabular}{p{58pt} p{78pt} p{45pt} p{45pt} p{45pt} p{45pt} p{36pt} p{39pt}<{\centering}}
\toprule
\multirow{2}{*}{Model (Region)} & \multirow{2}{*}{Noise Level} & \multicolumn{6}{c}{Evaluation Metrics} \\
\cmidrule(lr){3-8}
& & MAE $\downarrow$ & APE $\downarrow$ & PCK $\uparrow$ & FGD $\downarrow$ & BC $\uparrow$ & Diversity $\uparrow$ \\
\midrule
\multirow{4}{*}{F2G (Auditory)} & w/o noise & \textbf{0.603} & \textbf{0.874} & \textbf{0.451} & 25.1 & \textbf{0.628} & \textbf{163.7} \\
& mean=0, std=0.1 & 0.617 & 1.014 & 0.426 & 24.0 & 0.593 & 143.9 \\
& mean=0, std=0.5 & 0.728 & 1.251 & 0.266 & 21.4 & 0.418 & 106.2 \\
& mean=0, std=1.0 & 1.134 & 1.451 & 0.180 & \textbf{19.1} & 0.352 & 99.71 \\
\midrule
\multirow{4}{*}{F2G (Speech)} & w/o noise & \textbf{0.621} & 1.030 & \textbf{0.439} & 24.4 & \textbf{0.615} & \textbf{166.3} \\
& mean=0, std=0.1 & 0.624 & \textbf{0.908} & 0.420 & 24.2 & 0.585 & 141.5 \\
& mean=0, std=0.5 & 0.762 & 1.155 & 0.276 & 22.4 & 0.428 & 103.6 \\
& mean=0, std=1.0 & 1.041 & 1.616 & 0.188 & \textbf{18.6} & 0.347 & 98.25 \\
\midrule
\end{tabular}
\end{table*}

For all the compared methods, we set $N = 64$ frame clips and stride $M = 16$. The number of gesture keypoints was set to 49. For the potential dimensions of the fMRI signal, the dimension of the auditory region was considered to be 1,431, the dimension of the speech region was considered to be 498, and the 10,000 voxels through cross-validation approach with highest performance were selected as all regions. We fine-tuned the T2G model after the first stage training, and the loss balancing parameter $\lambda$ is set to 0.01. We employed the Adam optimizer with a learning rate of 1e - 4, and training was performed on dual NVIDIA GeForce RTX 3090 GPUs with a batch size of 32.

\subsection{Performance Comparison and Analysis}
\subsubsection{Comparison of using noise or fMRI as conditions}
To verify the necessity of using brain signals, rather than random signals, as generative conditions, we chose the noise condition for comparison. This condition conforms to the standard input format of the diffusion model and eliminates interference from other modalities. By fixing the text modality input (T2G generated results), we can independently compare the effects of the noise and fMRI conditions on generative performance, thereby isolating the semantic contribution specific to brain signals. The quantitative evaluations of the gestures generated by F2G using either Noise or fMRI as conditions are presented in Table \ref{tab:1}. The metrics were computed by comparing the generated gestures with the gestures produced by T2G (trained in the first stage). When using fMRI as the condition, F2G (fMRI) achieved a lower MAE and APE, and higher PCK and BC, indicating that these gestures correspond more closely with those from T2G. Moreover, the gestures generated by F2G (fMRI) exhibited higher Diversity, indicating that these gestures consist of more complex motions. In contrast, when using pure noise as the condition, FGD was lower, possibly because F2G (fMRI) integrates a new modality while F2G (Noise) is only related to the text modality as T2G.

\subsubsection{Comparison of using fMRI from different cortical regions as conditions}
In this experiment, we evaluated the impact of using fMRI data from different cortical regions as conditions. Given that the auditory cortex and language-related cortical regions are closely associated with the human language network \mycite{tang2023semantic}, which also participate in embodied cognition through semantic-action associations \mycite{binder2009semantic}, we focus on speech, auditory, and motor regions. The results are presented in Table \ref{tab:2}.
Using whole-brain fMRI data, F2G (All), as a control group is to verify the effectiveness of selecting function-specific brain regions. 
The best MAE, APE, and PCK values were observed for F2G (Auditory), with F2G (Speech) showing a slight superiority to F2G (Auditory) in terms of FGD. In contrast, F2G (All) yielded inferior results across most metrics. This suggests that for the task of co-speech gesture generation, the speech and auditory areas contain more semantic information that can represent gestures, while the brain activities from other areas may introduce more noise information. F2G (Speech) and F2G (Auditory) showed similar results overall, and combining speech and auditory areas did not further improve the results. This is possibly because these two areas share similar critical semantic information for gestures, rather than complementary information. Fig. \ref{fig:gesture} presents examples of the generated gestures using T2G and F2G. This aligns with the finding in Table \ref{tab:2}. It was found that incorporating the entire brain signal, which may include more noisy information, had a negative effect on the F2G model. F2G (All) produced gestures with minimal motion, which were the least similar to the gestures from T2G. On the other hand, utilizing signals from the auditory or speech regions resulted in more accurate gestures, indicating that both regions contain useful features highly relevant to gestures. To determine whether gesture generation primarily relies on semantics rather than motor cortex inputs, we conducted ablation studies utilizing motor cortex fMRI signals as the sole experimental variable. Using M1H signals led to a significant performance drop, further supporting our semantic-driven generation paradigm.


\subsubsection{Comparison across different noise levels}
To assess the impact of noise levels on fMRI signals, we conducted an evaluation by adding different levels of Gaussian noise to fMRI data from the auditory or speech areas. The noise injection experiment was carried out using a controlled variable method, performing a cross-comparison of noise responses between the auditory and language cortices. The mean of the noise ($\mu$ = 0) was fixed to avoid signal bias, while the standard deviation ($\sigma \in {0.1, 0.5, 1.0}$) was gradually increased to simulate different degrees of fMRI information loss. The experimental results are presented in Table \ref{tab:noise}. Lines 1-4 represent the auditory cortex, while lines 5-8 pertain to speech-related cortices. For example, when adding noise to the fMRI data of the auditory cortex with $\sigma=0.1$, minimal variations were observed in the results, indicating minimal effects on the gesture generation process. As expected, with an increase in the noise level, all the metrics exhibited a negative trend. Similar to the auditory cortex, we observed a comparable phenomenon for the speech area, where higher noise levels resulted in higher MAE, APE and lower PCK, FGD, BC, and Diversity. By comparing lines 2 vs 6, 3 vs 7, and 4 vs 8, we found that the noise had an equivalent impact on the model conditioned on the fMRI data from either the auditory cortex or speech-related cortices. Last but not least, the overall results demonstrate the ability of our F2G model on capturing the gesture semantics contained in brain activities.




\begin{table*}[tphb]
\centering
\caption{Human evaluation results for generated gestures, rated on a scale of 1-5.}
\label{tab:userstudy}
\setlength{\tabcolsep}{8pt}
    \begin{tabular}{p{74pt}ccccccc}
    \toprule
    Metrics & Naturalness & Smoothness & Human-likeness & Content Match & Understand-ability & Diversity & Overall rating \\
    \midrule
    Speech2Gesture \mycite{ginosar2019learning} & 1.45 & 2.59 & 3.24 & 3.21 & 3.31 & 1.46 & 2.42 \\
    JL2P \mycite{ahuja2019language2pose} & 1.38 & 3.32 & 2.35 & 3.73 & \textbf{3.58} & 3.24 & 3.00 \\
    Trimodal \mycite{yoon2020speech} & 3.54 & 3.23 & 2.87 & 3.12 & 3.01 & 3.12 & 3.30 \\
    DiffGesture \mycite{zhu2023taming} & 3.56 & \textbf{3.68} & 3.01 & 3.48 & 2.81 & 3.61 & 3.14 \\
    EMAGE \mycite{liu2024emage} & 3.22 & 3.55 & 2.95 & 3.72 & 3.29 & 3.58 & 3.38 \\
    fMRI2GES (Ours) & \textbf{3.71} & 3.19 & \textbf{3.30} & \textbf{3.87} & 3.41 & \textbf{4.04} & \textbf{3.57} \\
    \bottomrule
    \end{tabular}
\end{table*}

\begin{figure}[tbhp]
  \centering
    \includegraphics[width=\linewidth]{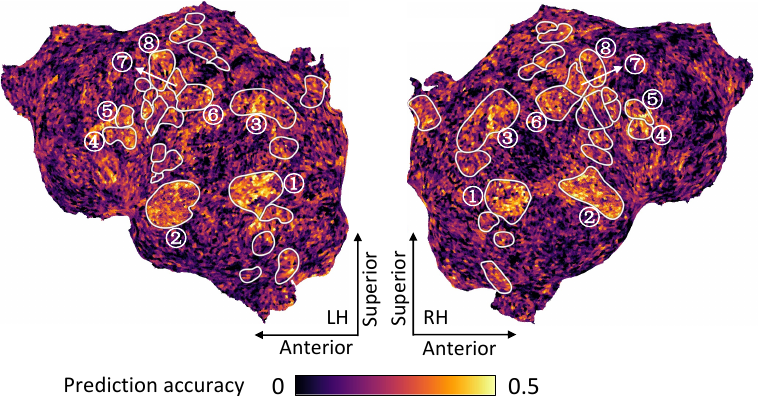}
    \vspace{-4mm}
    \label{fig:brain-a}
  \hfill
    \includegraphics[width=\linewidth]{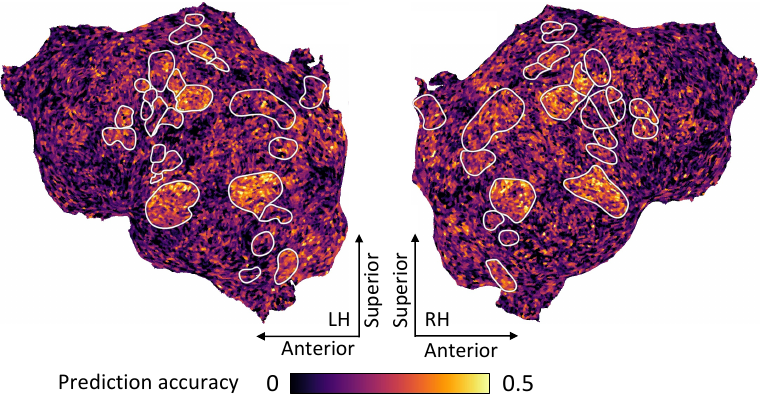}
    \vspace{-3mm}
    \label{fig:brain-b}
  \caption{Brain activity prediction performance assessed using Pearson's correlation coefficients, mapped onto the flattened cortical surface with the occipital areas at the center, covering both the left and right hemispheres. The prediction was conducted based on latent gesture representations extracted from the Unet in F2G, conditioned on fMRI data from the auditory cortex (upper) and speech-related cortices (lower). Brain areas: \ding{172} EBA (extrastriate body area), \ding{173} AC (auditory cortex), \ding{174} IPS (intraparietal sulcus), \ding{175} Broca (Broca's area), \ding{176} IFSFP (inferior frontal sulcus face patch), \ding{177} S1H (primary somatosensory cortex for hands), \ding{178} M1H (primary motor cortex for hands), \ding{179} FEF (frontal eye fields).}
  \vspace{-4mm}
  \label{fig:brain}
\end{figure}

\subsubsection{Predicting brain activities using latent gesture representations} To validate whether the learned gesture representation model captures the biological encoding mechanisms underlying brain gesture processing, we constructed an encoding model to map the latent representations of the F2G model to entire-brain voxel-wise brain activity. The latent gesture representations were extracted from the last layer of Unet in our F2G model, conditioned on fMRI data from either the auditory cortex (referred to as the AC condition in the following) or speech-related cortices (referred to as the SC condition in the following). Following \mycite{takagi2023high}, we utilized ridge regression as the mapping method and employed Pearson's correlation coefficients to measure the similarity between predicted and reference fMRI signals. The prediction accuracy is depicted in Fig. \ref{fig:brain}. Upon comparing the results, we observed that the AC condition achieved better overall prediction performance, yielding higher accuracy in the auditory cortex (AC) and speech-related cortices such as the Broca's area (Broca). Interestingly, we also observed high prediction results in several other brain areas. For instance, in the upper subfigure, the extrastriate body area (EBA), a component of the visual cortex, was prominently highlighted in bright yellow. Additionally, regions associated with motor and somatosensory functions, such as the intraparietal sulcus (IPS), the primary motor cortex for hand (M1H), and the primary somatosensory cortex for hand (S1H), displayed more pronounced results compared to the surrounding areas. These results demonstrate a mapping relationship between our representation model and brain regions implicated in gesture processing, indicating that using auditory and speech-related fMRI signals is both reasonable and feasible. Furthermore, the varying prediction accuracy across brain regions provides insights into specialized and multimodal brain functions, which could enhance future efforts to decode brain signals for applications such as brain-computer interfaces.

\subsubsection{Human evaluation}
To further evaluate the quality of the generated gestures, we conducted a comparative user study involving the proposed F2G and several other gesture generation methods. The selection of participants followed a double-blind principle, where participants were unaware of the correspondence between methods, and video sequences were presented randomly. The evaluation dimensions covered both low-level motion features (naturalness, etc.) and high-level semantic features (content matching, etc.). Since the model inputs vary, including speech or text, we synthesized speech audio with text as needed.
The comparison methods are: 
1) \textit{Speech2Gesture} \mycite{ginosar2019learning} uses spectrograms and employs an encoder-decoder neural network with adversarial training to efficiently map and accurately generate gestures corresponding to the speech.
2) \textit{JL2P} \mycite{ahuja2019language2pose} learns the mapping from language to animation in an end-to-end manner by jointly embedding language and pose.
3) \textit{Trimodal} \mycite{yoon2020speech} is an automatic gesture generation model based on multimodal context, reliably generating human-like gestures that match the content and rhythm of speech through an adversarial training scheme.
4) \textit{DiffGesture} \mycite{zhu2023taming} utilizes a diffusion-based framework for gesture generation, overcoming issues like mode collapse in GANs through the use of Diffusion Audio-Gesture Transformer and Gesture Stabilizer.
5) \textit{EMAGE} \mycite{liu2024emage} generates full-body gestures from audio and masked gestures, utilizing the BEAT2 dataset and a Masked Audio Gesture Transformer to enhance audio-to-gesture generation and masked gesture reconstruction through joint training, while improving gesture fidelity and diversity with adaptive audio feature fusion and VQ-VAE. We invited 60 volunteers, including 30 females and 30 males, aged between 18 and 35 years old, to participate in our study. The volunteers were asked to rate the generated gestures based on six criteria (naturalness, smoothness, human-likeness, content match, understandability, and diversity) using a scale of 1-5, as presented in Table \ref{tab:userstudy}.
The criterion of understandability measured the extent to which the gesture enhanced the understanding of the speech or text content. The overall rating was determined by calculating the average of the values for all the criteria. 
Our analysis revealed that the proposed F2G outperformed other methods in terms of naturalness, human-likeness, content match and diversity. The best smoothness was achieved by \textit{DiffGesture}, which utilized a gesture stabilizer to ensure temporal coherence. Overall, the participants generally agreed that the F2G model produced high-fidelity results in most aspects, thus verifying the effectiveness of the F2G model in generating natural and effective gestures.

\section{Conclusion}

In this work, we have addressed the challenge of decoding co-speech gestures from fMRI recordings, advancing the field of neuroscience and cognitive science. Our new approach, \textbf{fMRI2GES}, leverages the power of Dual Brain Decoding Alignment to train fMRI-to-gesture reconstruction networks on unpaired data, overcoming the limitations posed by the absence of paired \{brain, speech, gesture\} data.  Our findings not only demonstrate the efficacy of our proposed approach but also elaborate on the intricate relationship between fMRI signals from different Regions of Interest (ROIs) in the cortex and their impact on gesture generation results. This investigation contributes valuable insights into how the brain responds to external stimuli and enhances our understanding of the neural processes associated with co-speech gestures.

By bridging the gap between fMRI signals and expressive gestures, our approach not only pushes the boundaries of neuroscience research but also holds promise for applications in human-computer interaction, communication disorders, and other domains. We believe that our work paves the way for future advancements in brain decoding, with implications for both theoretical neuroscience and practical applications.



\bibliographystyle{IEEEtran}  
\bibliography{main}   

\begin{IEEEbiography}[{\includegraphics[width=1in,height=1.25in,clip,keepaspectratio]{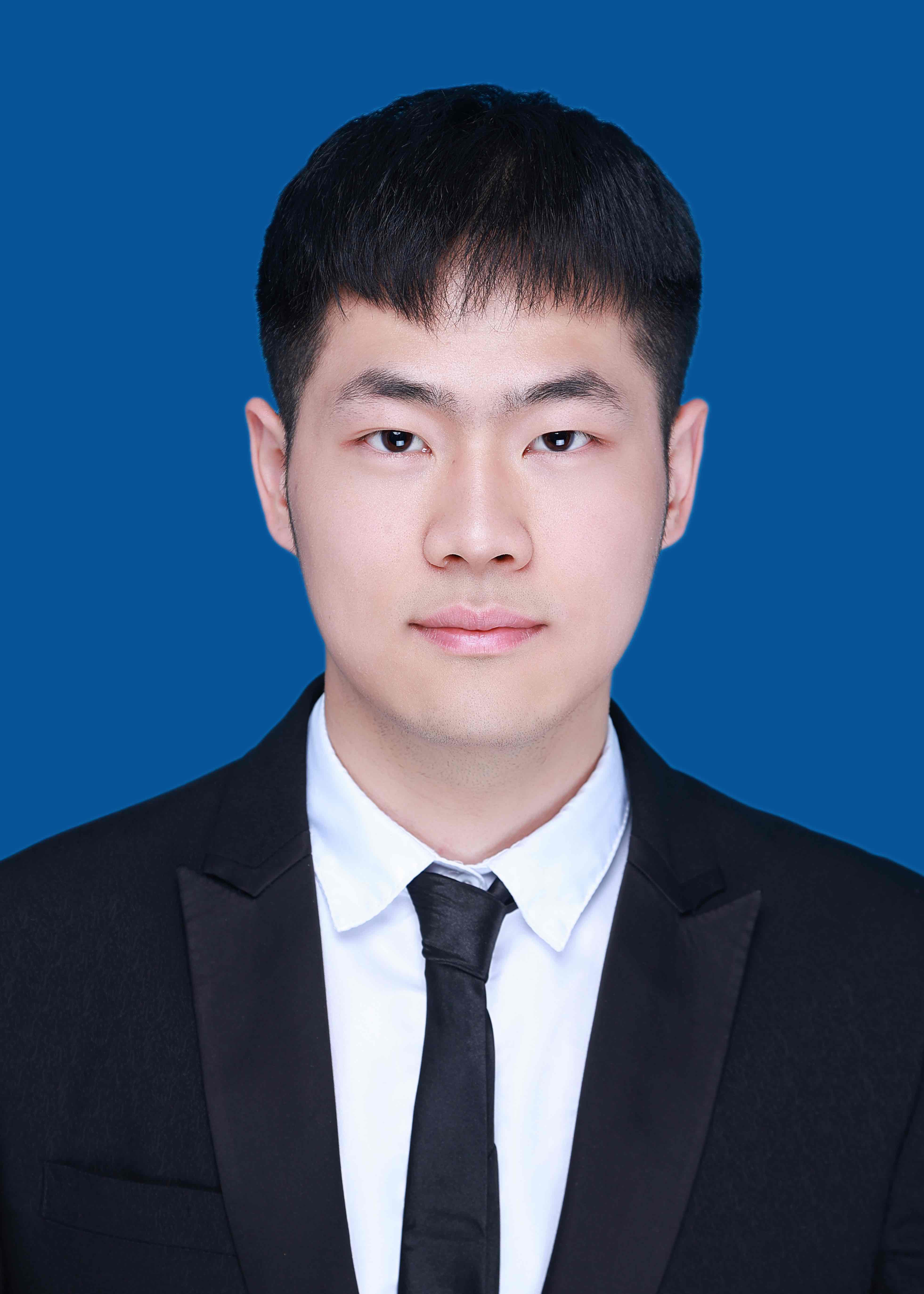}}]{Chunzheng Zhu} received his B.E. degree in 2022 from the School of Information Science and Engineering, Minzu University of China, Beijing, China, majoring in Data Science and Big Data. He is currently pursuing a Ph.D. in Computer Science and Technology at the College of Computer Science and Electronic Engineering, Hunan University, Changsha, China. His research interests include brain-computer interfaces, medical imaging, and self-supervised learning.
\end{IEEEbiography}
\begin{IEEEbiography}
[{\includegraphics[width=1in,height=1.25in,clip,keepaspectratio]{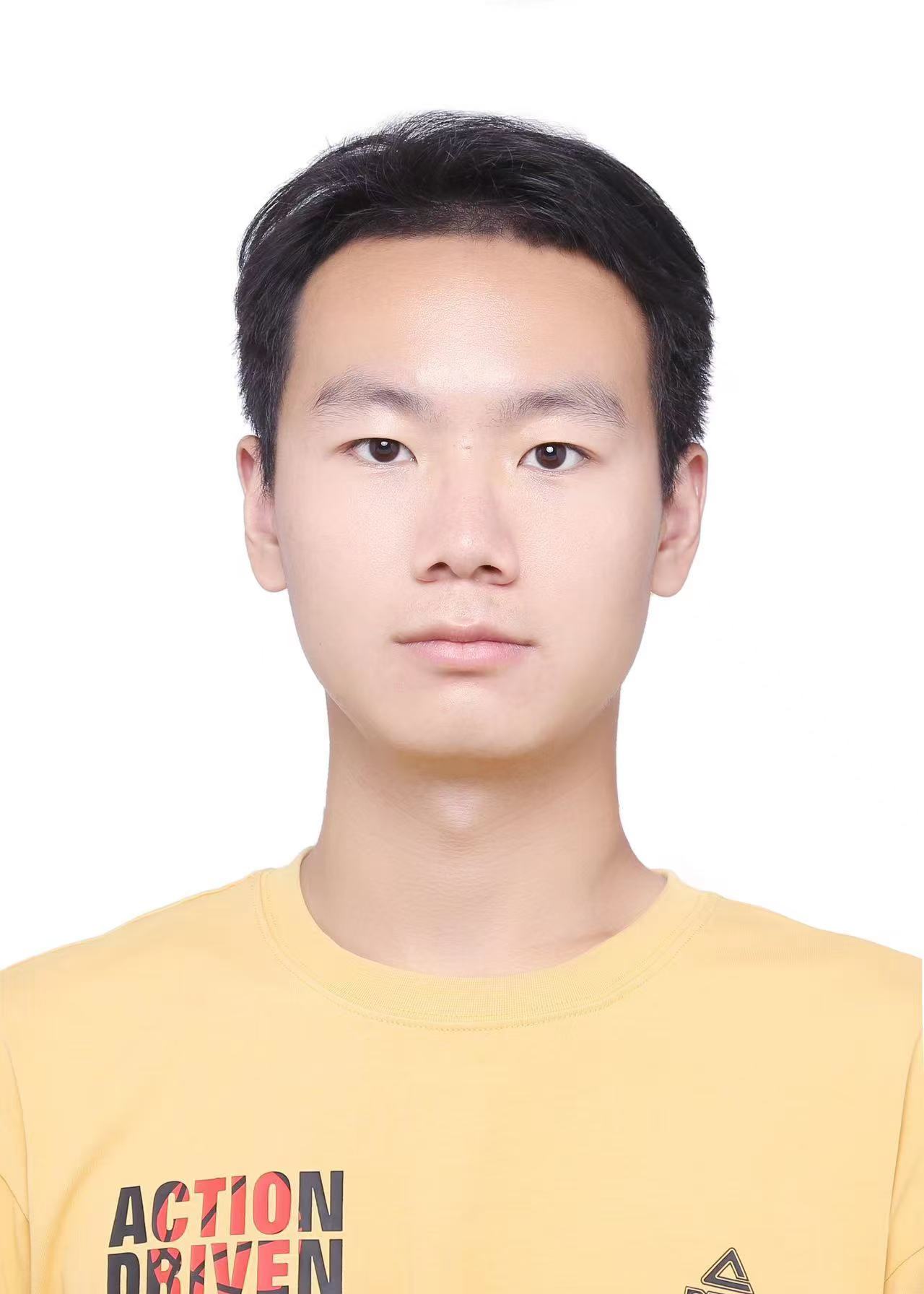}}]{JialinShao} received his B.E. degree from Zhengzhou University, Henan, China in 2023. He is pursuing an M.E. degree with the College of Computer Science and Electronic Engineering, Hunan University, Changsha, China. His research interests include ultrasound medical imaging, self-supervised learning, and few-shot learning.
\end{IEEEbiography}
\begin{IEEEbiography}
[{\includegraphics[width=1in,height=1.25in,clip,keepaspectratio]{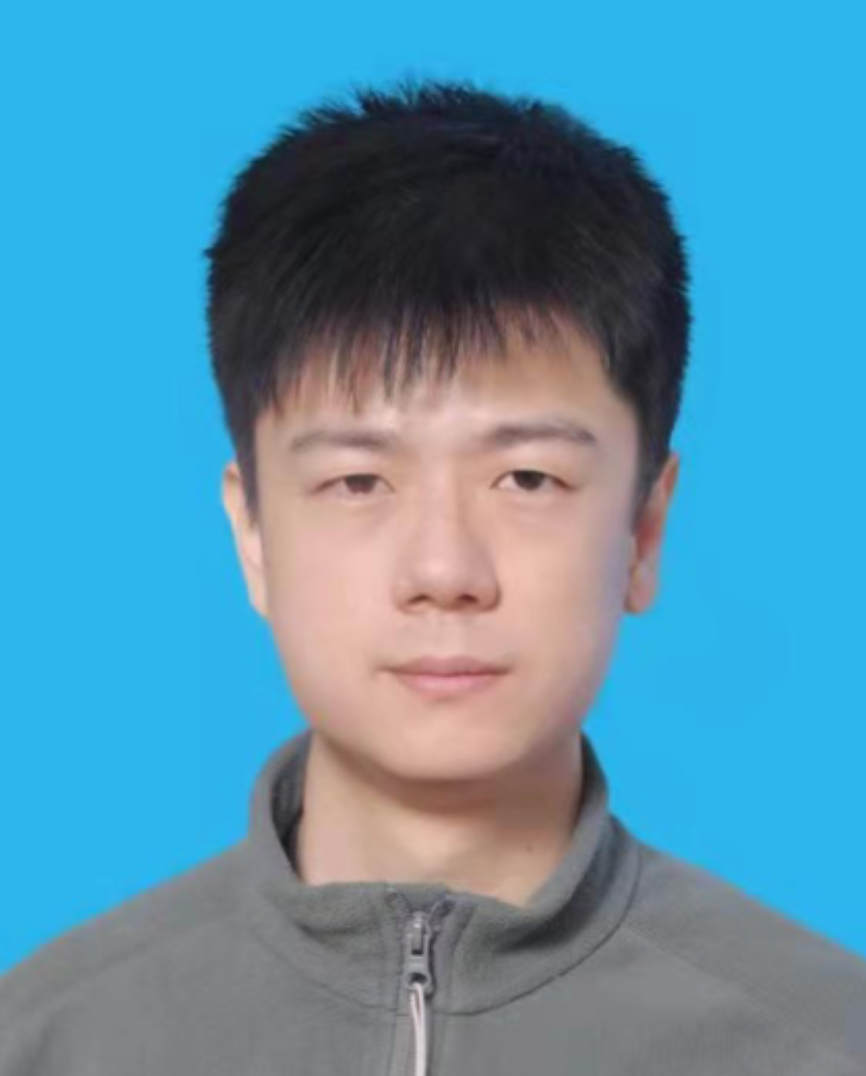}}]{JianxinLin} received the B.E. and Ph.D. degrees from the University of Science and Technology of China, Hefei, China, in 2015 and 2020, respectively. He is currently an Associate Professor with the School of Computer Science and Electronic Engineering, Hunan University, Changsha, China. He has authored or co-authored more than 20 papers in related conferences and journals. His research interests include image and video processing, synthesis, and understanding. He was the recipient of the top paper award at the ACM Multimedia 2022.
\end{IEEEbiography}
\begin{IEEEbiography}
[{\includegraphics[width=1in,height=1.25in,clip,keepaspectratio]{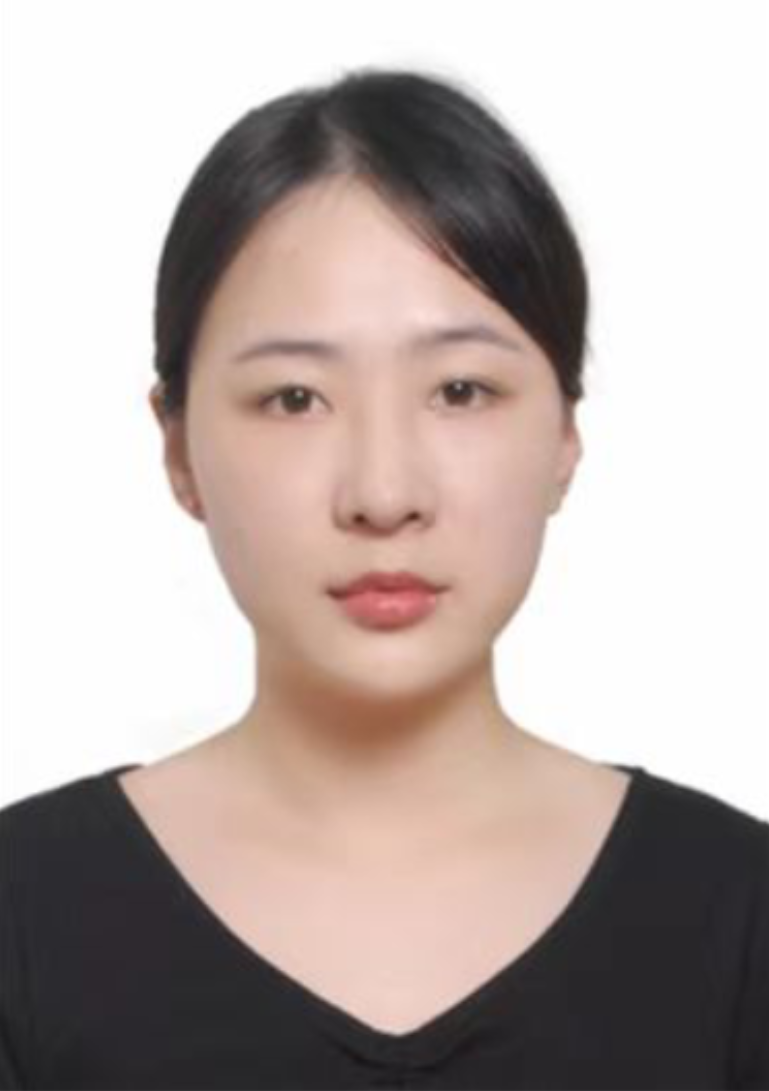}}]{YijunWang} received the B.E. and Ph.D. degrees
from the University of Science and Technology of China, Hefei, China, in 2014 and 2019, respectively. She is currently an Assistant Professor with the School of Computer Science and Electronic Engineering, Hunan University, Changsha, China. She has authored or co-authored more than 10 papers in related conferences and journals. Her research interests include multimedia understanding, natural language processing, and data mining.
\end{IEEEbiography}
\begin{IEEEbiography}
[{\includegraphics[width=1in,height=1.25in,clip,keepaspectratio]{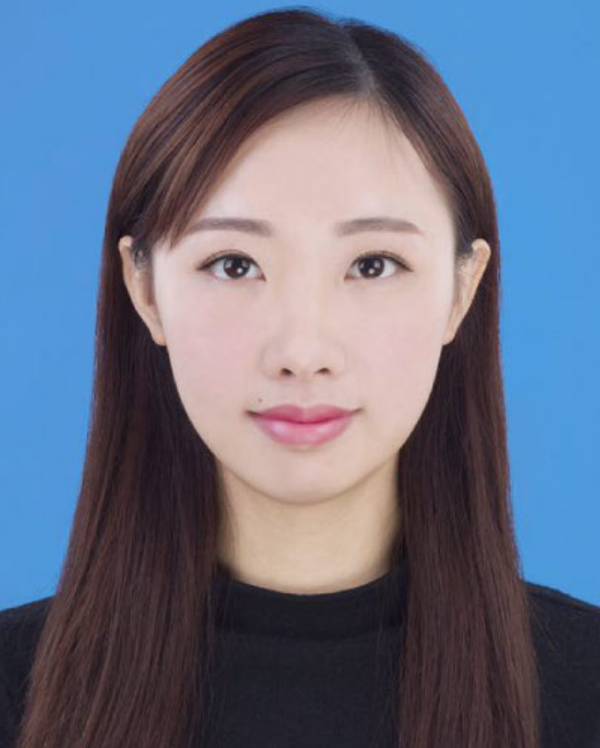}}]{JingWang} received the B.E. and Ph.D. degrees
from Nanjing University of Science and Technology, Nanjing, China, in 2015 and 2022, respectively. From 2019 to 2020, she was a Visiting Scholar with the University of Rochester, Rochester, NY, USA. She is currently a Post-Doctoral Fellow with the Department of Automation, Tsinghua University, Beijing, China. Her research interests include computer vision and multimedia analysis, with a focus on vision and language.
\end{IEEEbiography}
\begin{IEEEbiography}
[{\includegraphics[width=1in,height=1.25in,clip,keepaspectratio]{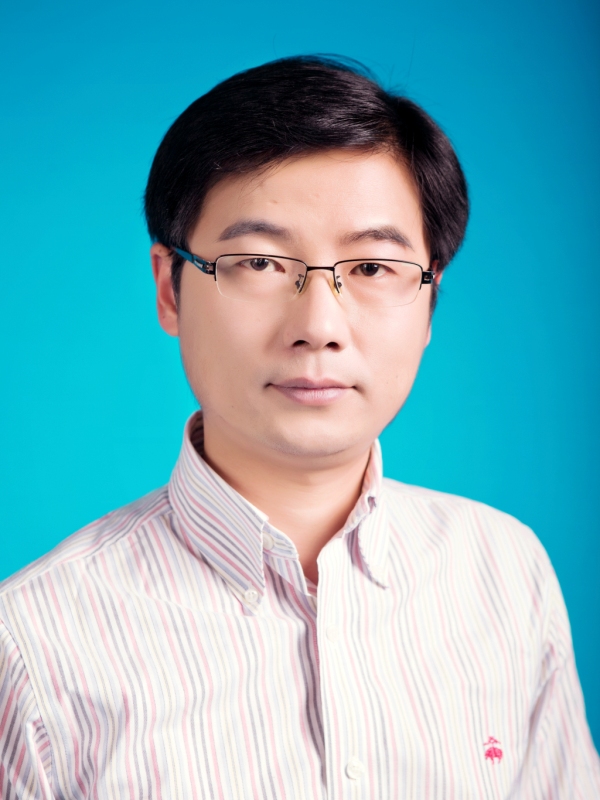}}]{JinhuiTang} (Senior Member, IEEE) received the B.E. and Ph.D. degrees from the University of Science and Technology of China, Hefei, China, in 2003 and 2008, respectively. He is currently a Professor with the Nanjing University of Science and Technology, Nanjing, China. He has authored more than 200 articles in toptier journals and conferences. His research interests include multimedia analysis and computer vision. Dr.Tang was a recipient of the Best Paper Awards in ACM MM 2007 and ACM MM Asia 2020, the Best Paper Runner-Up in ACM
MM 2015. He has served as an Associate Editor for the IEEE TNNLS, IEEE TKDE, IEEE TMM, and IEEE TCSVT. He is a Fellow of IAPR.

\end{IEEEbiography}
\begin{IEEEbiography}
[{\includegraphics[width=1in,height=1.25in,clip,keepaspectratio]{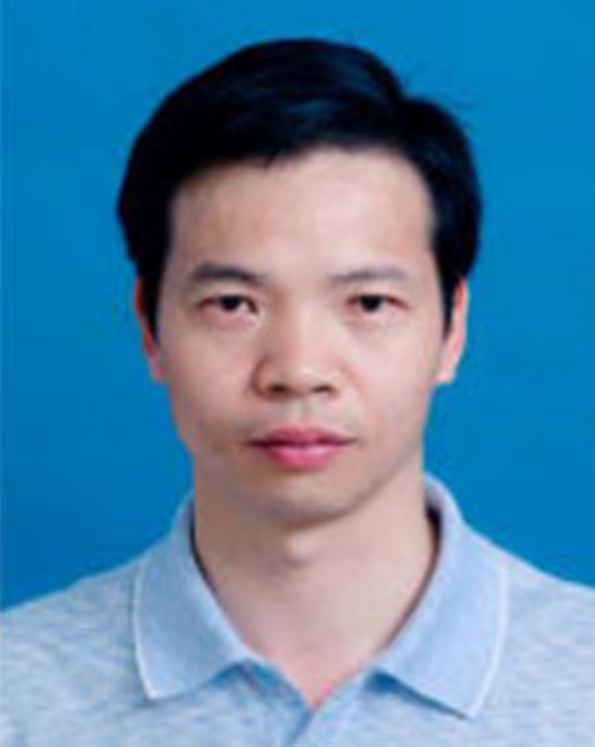}}]{KenliLi} (Senior Member, IEEE) received the Ph.D. degree in computer science from Huazhong University of Science and Technology, China, in 2003. He was a visiting scholar at the University of Illinois at Urbana-Champaign from 2004 to 2005. He is currently a Full Professor of computer science and technology at Hunan University and Associate Director of National Supercomputing Center in Changsha. His major research includes parallel computing, grid and cloud computing, and DNA computing. He has published more than 90 papers in international conferences and journals, such as IEEE-TC, IEEE-TPDS, JPDC, ICPP, and CCGrid.
\end{IEEEbiography}

\end{document}